# Automatic Micro-Expression Apex Frame Spotting using Local Binary Pattern from Six Intersection Planes

Vida Esmaeili
Faculty of Electrical and Computer Engineering
University of Tabriz
Tabriz, Iran
v.esmaeili@tabrizu.ac.ir

Mahmood Mohassel Feghhi
Faculty of Electrical and Computer Engineering
University of Tabriz
Tabriz, Iran
mohasselfeghhi@tabrizu.ac.ir

Seyed Omid Shahdi
Faculty of Electrical, Biomedical and Mechatronics
Qazvin Branch, Islamic Azad University
Qazvin, Iran
shahdi@qiau.ac.ir

*Abstract*— Facial expressions are one of the most effective ways for non-verbal communications, which can be expressed as the Micro-Expression (ME) in the high-stake situations. The MEs are involuntary, rapid, and, subtle, and they can reveal real human intentions. However, their feature extraction is very challenging due to their low intensity and very short duration. Although Local Binary Pattern from Three Orthogonal Plane (LBP-TOP) feature extractor is useful for the ME analysis, it does not consider essential information. To address this problem, we propose a new feature extractor called Local Binary Pattern from Six Intersection Planes (LBP-SIP*l*). This method extracts LBP code on six intersection planes, and then it combines them. Results show that the proposed method has superior performance in apex frame spotting automatically in comparison with the relevant methods on the CASME database. Simulation results show that, using the proposed method, the apex frame has been spotted in 43% of subjects in the CASME database, automatically. Also, the mean absolute error of 1.76 is achieved, using our novel proposed method.

## I. INTRODUCTION

In situations, where people are motivated to manipulate, conceal, or repress their true feelings, and there is no evidence to prove it, the Micro-Expressions (MEs) have received increasing attention, recently [1, 2]. These types of expressions have a wide range of applications, such as fair law enforcement, lie detection, identifying potentially dangerous persons, and more reliable diagnosis in the clinical psychology and psychotherapy [3, 4].

In fact, they are one of the universal emotions such as fear, happiness, surprise, disgust, sadness, anger, and contempt that happen when persons are deliberately trying to neutralize and hide them [4-7]. Unlike ordinary Facial Expressions (FEs), the MEs are a very brief, very rapid, subtle, and involuntary reaction with short time (from 0.04 to 0.33 seconds) and appears with low intensity [3]. These characteristics make it challenging, not only by the naked eye but also with tools in computer vision [8].

Haggard and Isaacs [9] discovered micro-momentary facial expressions as repressed emotions, while scanning motion picture psychotherapy films for the first time in 1966 [10]. Three years later, Ekman *et al.* [11] used the term MEs instead of the micro-momentary, analysing a video of a psychiatric patient interview, who tried to hide commit suicide from her psychiatrist [1, 12]. The patient pretended to be happy and optimistic throughout the recording, a fleeting look of anguish and despair that lasted for merely two frames (0.08s) was found, when the tape was examined curiously in the slow motion. Meanwhile, a brief anguish was quickly replaced by a smile, when reviewing the video frame by frame. Then, she confessed to lying to her doctor in her another counselling session [1, 13]. Thus, Ekman *et al.* found a relationship between the ME and the lie or deception.

Hence, there have been some efforts for the ME detection and recognition. Among presented approaches, Local Binary Pattern (LBP)-based methods are more suitable for ME feature extraction [2, 14]. These methods are appearance-based methods that extract facial features from skin texture (such as wrinkles on the chin and around the eyes, lines around the mouth) [2, 8]. The LBP on Three Orthogonal Planes (LBP-TOP), introduced in [15], is used to extract the MEs in [16]. Since the LBP-TOP well-performed under subtle facial motions, Ruiz-Hernandez *et al.* [14] employed re-parameterization of second-order Gaussian jet on it to create more robust and reliable histograms. The robustness to some spatial transformations and lighting changes are two benefits of this approach, although it generates high-dimensional feature sets with redundant information, and has computational complexity.

For reducing redundant information, Wang *et al.* [17] proposed the LBP with Six Intersection Points (LBP-SIP). In their method, the neighbouring points is just considered once in comparison with the LBP-TOP, which are computed twice. Not only it provides a more compact and lightweight representation but also reduces the computational complexity, and the time for feature extraction. Nevertheless, there was no significant difference in terms of the accuracy between these two approaches, namely LBP-TOP and LBP-SIP [17].

Afterward, Huang et al. [8] mentioned two problems of the LBP-TOP: 1) Extracting motion and appearance features from the sign-based difference between two pixels, without considering orientation and magnitude components, 2) Using classical pattern, which is not proper for local structures.

Therefore, they proposed the Spatio-Temporal Completed Local Quantization Patterns (STCLQP) [8] as a flexible encoding algorithm, which extracts sign, magnitude and orientation components for improving the performance of the ME recognition. The STCLQP, same as other methods, extracts the components from only three orthogonal planes without considering other planes.

After feature extraction, the apex frame spotting is important for the ME spotting and recognition, too. The apex frame contains maximum facial muscle movements and peak intensity of the ME throughout the video sequence. Thus, we can separate the MEs from the FEs only with computing the time of neutral face frame and the apex frame throughout the ME image sequence. Also, the apex frame can be used for FE classification alone.

Since the MEs are short and subtle in the time and intensity, the apex frame finding is challenging. Although researchers have tried to reduce the mean distance of spotted apex frame from the ground-truth apex, they could not determine the apex frame number exactly. Recently, Liong et al. [1] have obtained the apex frame with a deviation of one frame from the ground-truth apex in a sample.

Also, Ma et al. [18] have presented the Region Histogram of Oriented Optical Flow (RHOOF) to spot the apex frame automatically. They achieved improvements of 19.04% in comparison with the previous works in the CASME database. However, the Mean Absolute Error (MAE) can be decreased by considering the necessary information. Furthermore, the subtle changes and the maximum motions can appear in the image sequences, the temporal planes could be essential to detect them. In fact, these planes extract temporal features.

For this reason, in this paper, we propose the Local Binary Pattern from Six Intersection Planes (LBP-SIP$l$), where the temporal features are attained using six planes in multiple orientations. The framework for the proposed LBP-SIP$l$ is shown in Fig. 1. Initially, the LBP has exploited on the six planes. Then, all obtained histograms have been concatenated together to make a single histogram.

The contributions of this paper are as follows:
- A novel feature extractor called LBP-SIP$l$ is proposed to reveal very subtle motions in any directions.
- The exact apex frame is automatically spotted in several samples.
- The MAE and the Standard Error (SE) of the proposed method are reduced compared to the previous methods.

The remaining parts of the paper are organized as follows: In Section II, we express our proposed LBP-SIP$l$. Afterward, experimental results are given in Section III. Finally, we conclude the paper and suggest future directions in Section IV.

## II. THE PROPOSED METHOD

In this section, initially, we glimpse into the LBP-TOP, which is the base of our proposed method. Then, the proposed LBP-SIP$l$ will be presented in detail.

### A. Review of LBP-TOP

The LBP-TOP [15] contains three orthogonal planes (XZ, YZ, XY), intersected each other in the central pixel (Fig. 2). In this method, the sequential dynamic texture images are considered as the XY planes. Their horizontal and vertical pixels with corresponding locations are placed into the XZ and YZ planes, respectively. Then, the LBP code [19] is computed on each the orthogonal plane. Finally, the obtained histograms from three planes are concatenated into a single histogram. The mentioned approach generally extracts the motion and the appearance features from three planes, while it does not consider other essential information.

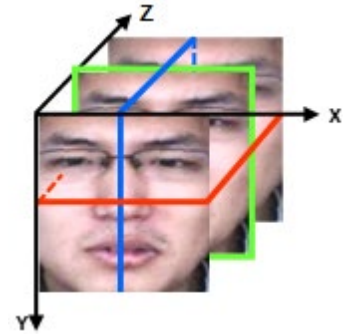

Fig. 2. Three planes in the LBP-TOP.

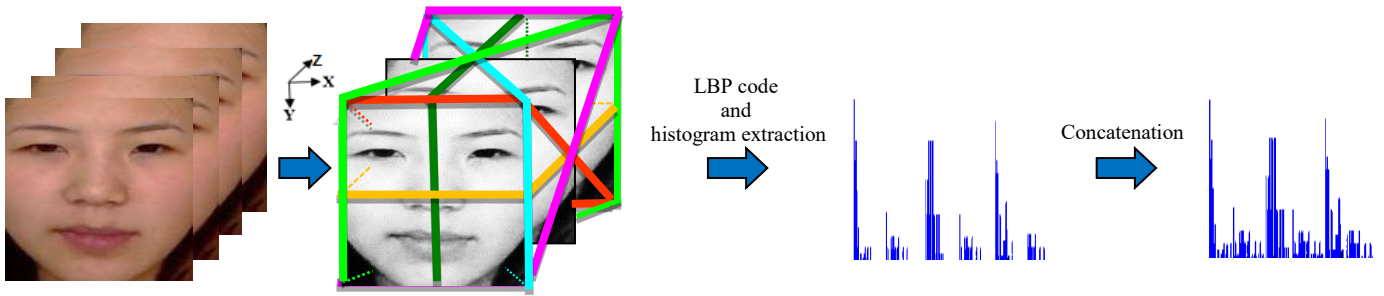

Fig. 1. Framework for the proposed LBP-SIP$l$ method. (a) Face alignment (b) Six planes of LBP-SIP$l$ (c) LBP extraction from each plane and their corresponding histograms (d) Concatenating to make a single histogram.

## B. The proposed LBP-SIP*l*

To address the limitations of the LBP-TOP, we propose the LBP-SIP*l*, which consists of the six planes. In this method, we extract more meaningful information by combining the histograms of the six temporal planes.

The six planes are including two temporal planes of the LBP-TOP and four new planes. The four new planes show the motions of the facial muscle simultaneously in both horizontal and vertical directions. These six planes intersect in the central pixel.

Now, consider a sequence of images, and suppose that they make a three-dimensional volume ($X$, $Y$, $Z$). If $g_{tc,c}$ is corresponding to the gray value of the central pixel of current frame, its coordinates are ($x_c$, $y_c$, $z_c$). Similarly, the $g_{tc-L,c}$ and $g_{tc+L,c}$ are the gray value of the central pixel of the previous and the posterior frames, respectively.

The coordinates of the $g_{tc,c}$ are given by ($x_c+R_x cos(2\pi n_p/N_P)$, $y_c-r_y sin(2\pi n_p/N_P)$, $z_c$), where $n_p \in 0, ..., N_P-1$ is the local neighbouring point around the central pixel. The $N_P$ and $r$ are the number of local neighbouring points and the radius of the circle (distance from central pixel to the neighbour points on the circle), respectively. The average of nearest pixels are computed to estimate a new pixel value, whenever the values of the neighbors do not fall exactly on a circle. Therefore, the coordinates of the neighboring points around the central pixel in planes number 1, 2, …, 6, (Fig. 3) are given by:

$(x_c + r_x cos(2\pi n_p/N_P), y_c, z_c + r_z sin(2\pi n_p/N_P))$    (1)

$(x_c, y_c - r_y sin(2\pi n_p/N_P), z_c + r_z cos(2\pi n_p/N_P))$    (2)

$(x_c + r_x cos(2\pi n_p/N_P), y_c - r_y sin(2\pi n_p/N_P), z_c + r_z cos(2\pi n_p/N_P))$ (3)

$(x_c + r_x cos(2\pi n_p/N_P), y_c - r_y sin(2\pi n_p/N_P), z_c - r_z cos(2\pi n_p/N_P))$ (4)

$(x_c - r_x cos(2\pi n_p/N_P), y_c - r_y sin(2\pi n_p/N_P), z_c - r_z sin(2\pi n_p/N_P))$ (5)

$(x_c + r_x cos(2\pi n_p/N_P), y_c - r_y sin(2\pi n_p/N_P), z_c + r_z sin(2\pi n_p/N_P))$ (6)

The LBP code is computed for each plane (refer to [15] for more detail). Subsequently, the histogram of the LBP-SIP*l* can be defined as:

$$hist_{j,k} = \sum_{x,y,z} I\{f_j(x,y,z) = j\}, \quad j = 0,...,n_k-1; \; k = 0,1,2,...,5 \quad (7)$$

where $f_j(x,y,z)$ is the LBP code of central pixel $(x,y,z)$ in the k-th plane, $n_k$ is the number of different labels processed by the LBP operator in the k-th plane and

$$I(B) = \begin{cases} 1, & \text{if B is true}; \\ 0, & \text{if B is false}. \end{cases} \quad (8)$$

Finally, the histograms are normalized using:

$$N_{j,k} = hist_{j,k} / \sum_{m=0}^{n_k-1} hist_{m,k} \quad (9)$$

## III. EXPERIMENTS AND RESULTS

The details of the implementation of our proposed method are given in this section. The performance of our proposed method and its comparison with other methods are also provided. It is noteworthy that the implementation of the proposed method is performed with MATLAB R2017a using a 3.5GHZ Intel Core i7 Duo processor and 8GB RAM on board.

## A. The database

We perform our experiments on the Chinese Academy of Sciences Micro Expressions (CASME) [4] database. An example of a part of the frame sequence from the CASME database is shown in Fig. 4. This database is a collection of 195 videos of spontaneous and dynamic MEs at a frame rate of 60 fps. The videos filmed using two BenQ M31 and Point Grey GRAS-03K2C cameras with the spatial resolutions of 1280×720 and 640×480 pixels [4]. Then, videos have been converted to the sequential images. Some of the advantages of the aforementioned database are as follows [4]:

- Having Action Unit (AU) labels
- Having neutral faces before and after each ME clip
- Labelling clips on onset (the occurrence of ME), apex and offset (the disappearance of ME) frames
- Being relatively "pure and clear" without noises such as irrelevant facial and head movements
- Classifying the emotions into seven classes.

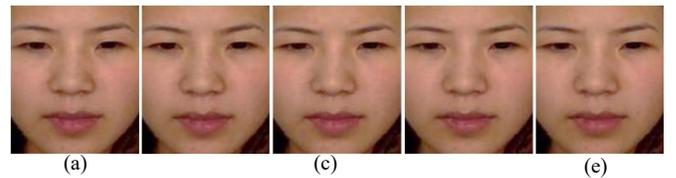

Fig. 4. An example for a part of the frame sequence from the CASME [4] database including onset frame (a), apex frame (c), and offset frame (e).

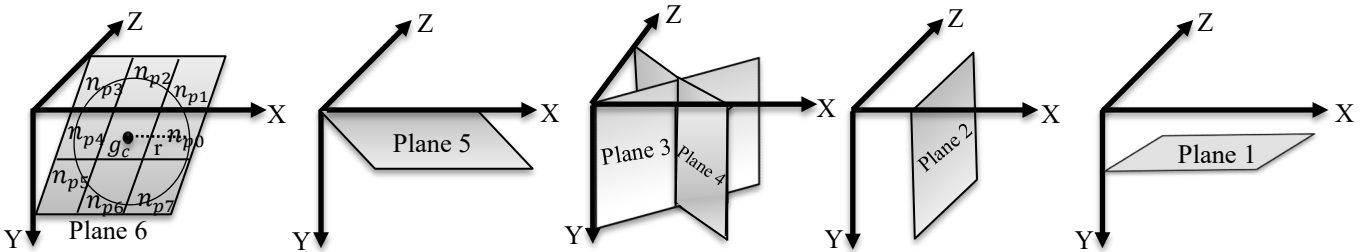

Fig. 3. Six planes in the LBP-SIP*l*.

## B. Feature extraction using the proposed LBP-SIPl

The sequential colour images of each sample from the CASME database are received. These images are then converted to grayscale. The pre-processing techniques are applied to them. The same size image sequences are taken as a three-dimensional array.

We assume circles with unit radius along X, Y, and Z axes. The number of the neighbouring points in the planes can be 4, 8, 16, and 24. The LBP-SIP*l* histogram as a feature vector is obtained as a matrix with the size $6 \times 2^{N_P}$, in which 6 is the number of planes, and $N_P$ is the number of neighbouring points. We consider NP=8 neighbouring points in this paper. The elapsed time for each the LBP-SIP*l* histogram is 0.8 seconds based on the simulation run time.

## C. Apex frame spotting

For apex frame spotting, the obtained histograms should be compared with the neutral frame histogram. So that the frame which contains the most shadow changes and wrinkles, is determined as the apex frame.

To find the difference between two frames, we compute the sum of squared differences of two histograms as follows:

$$\sum_{i=0}^{n} (h_1 - h_i)^2 \quad (10)$$

where $h_1$ is the first frame (neutral face) histogram, and $h_i$ is the current frame histogram in the image sequence with $n$ frame.

## D. Results and Discussion

A sample of the apex frame spotting using each plane is shown in Fig. 5. In this sample (i.e., sub08-EP12-2-2 in CASME [4] database), the ground-truth apex is 12. Also, we put six planes in one figure, shown in Fig. 6, to find apex frame using the LBP-SIP*l*. Since each plane shows the changes and motions in located direction; thus, the frame with the most feature differences in every direction is the apex. As can be seen, the spotted apex frame is exactly 12. In Fig. 5 and Fig. 6, vertical axis shows the Feature Difference (FD) of each frame from first frame, and the horizontal axis is the number of frames.

The percentage of the apex frame finding using each six planes of the proposed method and total of them are shown in Fig. 7. According to the results, the percentage of the apex frame spotting by the proposed method is 43%. It means that in 43% of all CASME samples, spotted apex is the ground-truth apex.

Plane 3, plane 4, plane 5, and plane 6 display both of changes in the row and the column, while plane 2 shows one column of the pixels changing in the time. Since facial muscle movements are less in vertical direction, the plane 2 does not have a good performance. In addition, the temporal planes are more important than the XY plane, as the XY plane only shows the appearance changes and it does not contain the motion transition information.

Finally, we compute the MAE and the SE for comparing our method with other methods. The MAE estimates the number frames that the spotted apex frame is off the ground-truth. It is defined for $N$ sample size as [18]:

$$MAE = (1/N) \sum_{i=1}^{N} |e_i| \quad (11)$$

where $e$ is deviation of the spotted apex frame from the ground-truth. The SE is the standard deviation of the sample means distribution. It is defined as [18]:

$$SE = \sigma / \sqrt{N} \quad (12)$$

where $\sigma$ is the sample deviation, and $N$ is the sample size.

The experimental results are illustrated in Table I. As can be seen, the MAE and the SE achieved by our method are smaller than the MAE and the SE achieved by the LBP, RHOOF [18] and LBP-TOP methods on the CASME database. They verify significant error reduction in our work compared to the aforementioned previous methods.

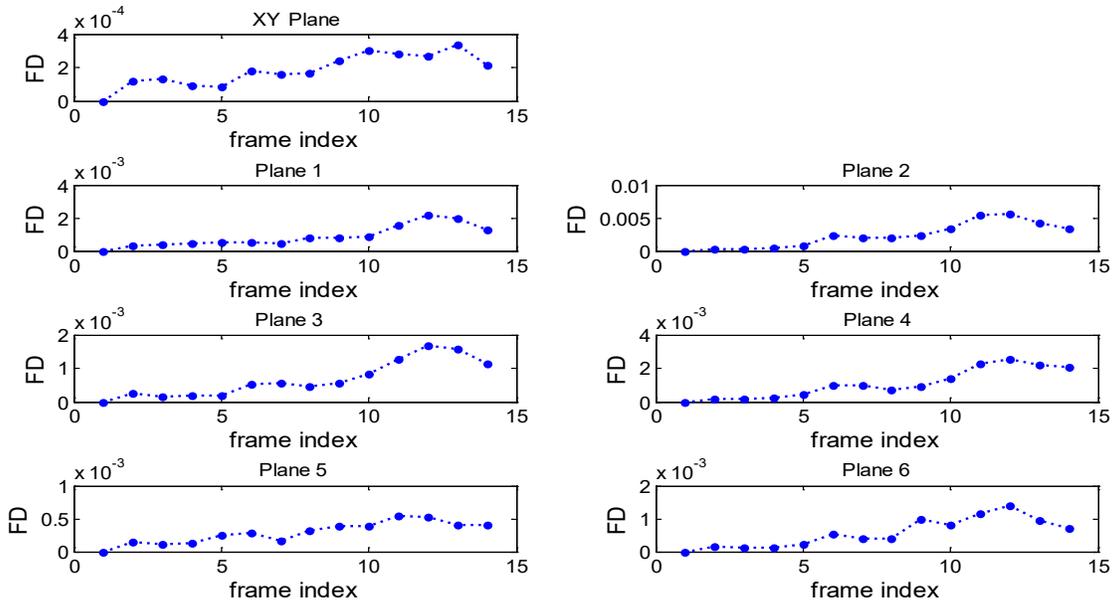

Fig. 5. The results of the apex frame finding in six planes of the proposed method and the XY plane.

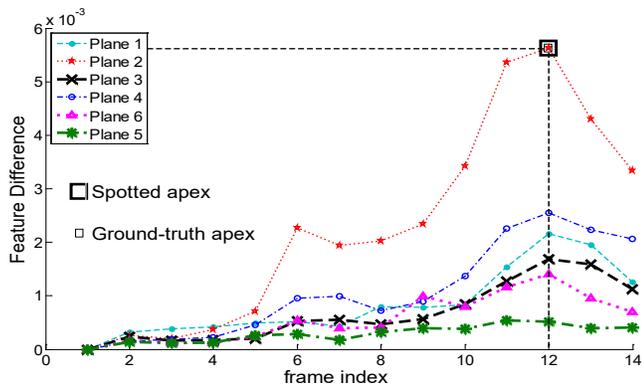

Fig. 6. Apex frame spotting using the LBP-SIP*l* method.

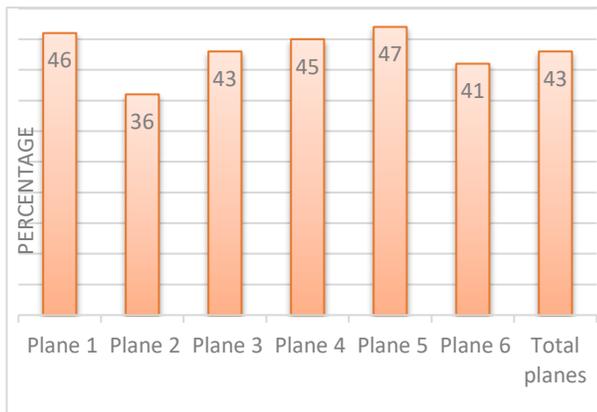

Fig. 7. The percentage as apex frame spotting using the LBP-SIP*l* method (total planes) and each plane of that on the CASME database.

TABLE I. THE RESULTS OF THE APEX FRAME SPOTTING USING THE LBP-SIP*l* METHOD ON THE CASME DATABASE.

| Method | MAE | SE |
| --- | --- | --- |
| LBP (BS-RoIs) [20] | 5.20 [18] | 0.58 [18] |
| RHOOF [18] | 3.60 [18] | 0.35 [18] |
| LBP-TOP | 2.54 | 0.23 |
| **LBP-SIP*l*** | **1.76** | **0.12** |

## IV. CONCLUSION

The MEs could reveal real human intention, since they are uncontrollable and spontaneous. These characteristics give an exceptional opportunity to be used in the wide range of applications from judgment court to psychology research centers. However, it could only be fruitful, when the feature descriptor is able to extract all short duration and low intensity changes. For this reason, in this paper, we proposed a novel method, named as LBP-SIP*l*, comprising six planes, which could extract subtle facial movements. According to our numerical experiments, the LBP-SIP*l* has superior performance in apex frame spotting comparing to the related existing methods. The numerical results showed that the apex frame was found in 43% of all CASME samples automatically. For further researches in the future, the LBP-SIP*l* could be applied in situations, where the detection of small variations are required.